\documentclass[journal,10pt]{IEEEtran}
\usepackage[T1]{fontenc}
\usepackage{amsmath,amssymb,booktabs,tabularx,array,graphicx,xspace,cite}
\usepackage[hidelinks]{hyperref}
\newcommand{\ASR}{\operatorname{ASR}}
\begin{document}
\title{Learning to Coordinate: LLM-Guided Workflow Evolution for Token-Efficient Red Teaming}
\author{Xiaoyang Xing, Wenyu Dai, Xiaoliang Gong, Asoke K Nandi, and Anthony Cohn%
\thanks{Xiaoyang Xing, Wenyu Dai, and Xiaoliang Gong are with the School of Computer Science and Technology, Tongji University, Shanghai, China (emails \href{mailto:augustusper@tongji.edu.cn}{\mbox{augustusper@tongji.edu.cn}}, \href{mailto:2451688@tongji.edu.cn}{\mbox{2451688@tongji.edu.cn}}, and \href{mailto:gxllshsh@tongji.edu.cn}{\mbox{gxllshsh@tongji.edu.cn}}).}%
\thanks{Asoke K Nandi is with Electronic and Electrical Engineering, Brunel University of London, Uxbridge, UK (email \href{mailto:Asoke.Nandi@brunel.ac.uk}{\mbox{Asoke.Nandi@brunel.ac.uk}}).}%
\thanks{Anthony Cohn is with the School of Computer Science, University of Leeds, Leeds, UK (email \href{mailto:A.G.Cohn@leeds.ac.uk}{\mbox{A.G.Cohn@leeds.ac.uk}}).}%
\thanks{For correspondence, email \href{mailto:augustusper@tongji.edu.cn}{\mbox{augustusper@tongji.edu.cn}}.}}
\hypersetup{
  pdftitle={Learning to Coordinate: LLM-Guided Workflow Evolution for Token-Efficient Red Teaming},
  pdfauthor={Xiaoyang Xing, Wenyu Dai, Xiaoliang Gong, Asoke K Nandi, Anthony Cohn}
}
\maketitle
\begin{abstract}
Automated red teaming is essential for identifying safety failures in large language models (LLMs), but its practical scalability depends on the resources required to discover those failures. Multi-agent systems can combine complementary capabilities, yet repeated planning, redundant exchanges, and ineffective information routing can make their coordination expensive. Moreover, a target model's response indicates the outcome of a test without identifying which collaboration decision should change. We present Learning to Coordinate (L2C), an LLM-guided framework that evolves an executable multi-agent workflow for each red-teaming case. L2C represents agent responsibilities, execution dependencies, and input bindings in a versioned graph, and associates each graph with records of the information actually delivered and produced. A controller trained through supervised fine-tuning integrates these records with target responses, evaluator verdicts, and remaining budget to retain the workflow, refine instructions, restructure collaboration, or lock further editing. This design connects feedback to explicit changes in subsequent execution while allowing a useful organization to continue without repeated controller intervention. The evaluation protocol covers HarmBench with three execution backbones and two target models, together with component ablations, dataset transfer, and evaluator checks. Across the six executor--target configurations, L2C improves ASR over EvoSynth by 1.4--4.5 percentage points while reducing mean online tokens by 35.6--37.3\%. L2C connects adaptive collaboration to the resource requirements of automated security evaluation.

\end{abstract}
\section{Introduction}
\label{sec:introduction}
Large language models (LLMs) increasingly mediate access to information, generate software, and support decisions in user-facing applications. Adversarial inputs can induce behavior that conflicts with these systems' intended safeguards. Studies of in-the-wild jailbreaks document sustained efforts to discover and share such inputs~\cite{dan}. Red teaming investigates these failures through human testing~\cite{ganguli} and automated test generation~\cite{perez}. As models and deployments change, evaluations must be repeatable at a manageable inference cost.

\suppressfloats[t]
\begin{figure}[t]
\centering
\includegraphics[width=\columnwidth]{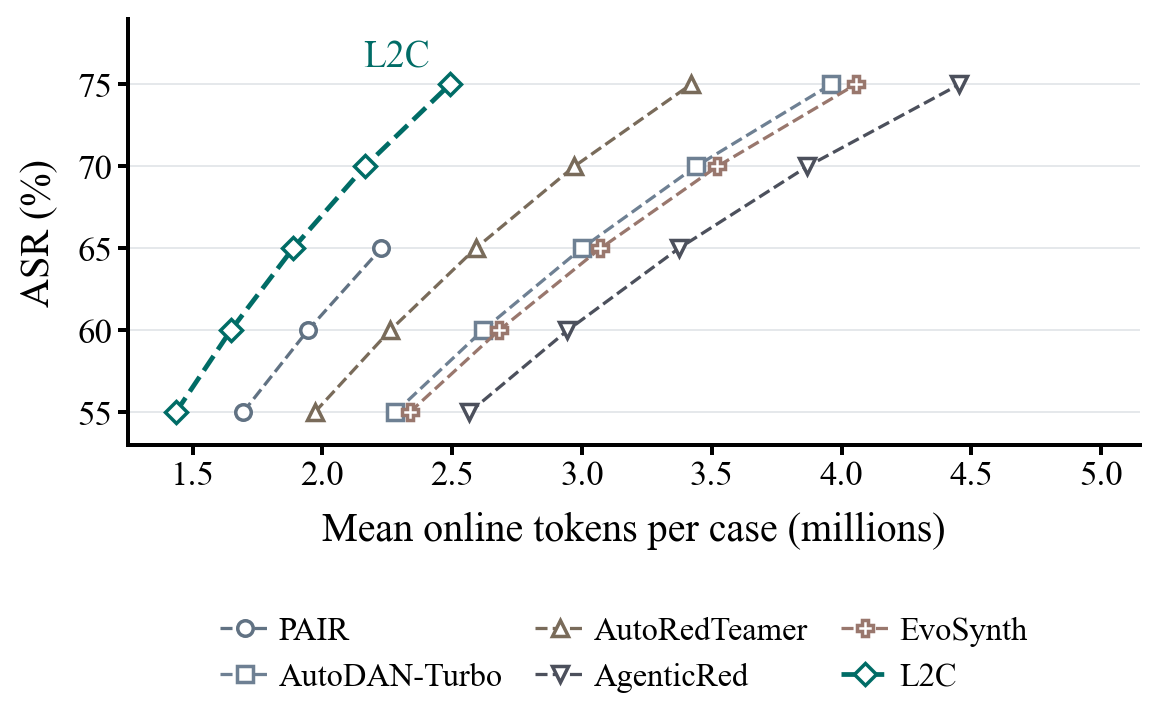}
\caption{Estimated success--token comparison for the Mixtral-8x7B executor and Llama-3.1-70B-Instruct target. Curves illustrate token costs at matched attack success rates.}
\label{fig:matched-asr}
\end{figure}

Automated red teaming has progressed from optimizing individual prompts~\cite{gcg,autodan} to feedback-guided refinement~\cite{pair} and structured search~\cite{tap}. More recent methods treat the testing system itself as an optimization target. AgenticRed evolves reusable agentic systems~\cite{agenticred}, and EvoSynth searches over executable attack programs with feedback-driven repair~\cite{evosynth}. Once a workflow is in use, the organization that suits one case can duplicate work or omit useful information in another. The controller must decide whether to retain the workflow, revise an agent's instructions, or change how agents collaborate.

This decision requires evidence about both the outcome and the execution that produced it. A negative verdict establishes that the target response failed the evaluation criterion. It offers little guidance about which agent or dependency needs attention. For example, an agent can produce a useful review that never reaches the agent preparing the submitted input. The execution record exposes that omission by showing what the receiving agent actually saw. The next pass can deliver the review through an explicit input binding. Because outcomes are stochastic, the effect of the change still has to be measured across cases and runs.

Diagnosing and revising a workflow consume the same token budget as executing it. A system can reduce target queries and still spend more tokens on planning, intermediate exchanges, and assessment. Figure~\ref{fig:matched-asr} illustrates the comparison at matched success rates. \emph{How can an LLM use execution feedback to adapt collaboration for an individual red-teaming case and achieve comparable or higher success with fewer total online tokens?}

We introduce \textbf{Learning to Coordinate (L2C)}, which connects a learned controller to an executable, versioned workflow graph. Nodes specify responsibilities and interfaces, edges constrain execution order, and bindings determine the evidence delivered to each node. After a pass, the controller uses the graph, execution records, target outcome, and remaining budget to decide what to change. It can also lock the graph so that subsequent passes use fresh case feedback without further editing calls. Supervised fine-tuning combines valid action demonstrations with decisions selected from execution outcomes. Controller parameters stay fixed during evaluation. Each case maintains its own graph and execution history.

GRAFT~\cite{graft} and InFlowOp~\cite{inflowop} explore runtime adaptation, and AOrchestra studies learned delegation~\cite{aorchestra}. L2C addresses the red-teaming setting by connecting verdicts about the target response to records of agent collaboration and charging the decisions to the same budget as execution.

We make three contributions.
\begin{itemize}
\item We represent case-specific workflow evolution through graph edits to responsibilities, dependencies, and input bindings. Versioned traces record the inputs each agent received and the outputs it produced.
\item We train a controller to retain, refine, restructure, or lock the workflow using execution evidence and remaining budget. The action space distinguishes continued execution from continued editing.
\item We evaluate attack success and total online resource use on HarmBench~\cite{harmbench} across three execution backbones and two targets. Matched-success comparisons and component ablations examine when adapting collaboration reduces token expenditure.
\end{itemize}

\section{Related Work}
\label{sec:related}
\subsection{Automated and Agentic Red Teaming}
Automated red teaming explores adversarial inputs through gradient optimization~\cite{gcg}, genetic search~\cite{autodan}, template mutation~\cite{gptfuzzer}, iterative refinement~\cite{pair}, and branching search~\cite{tap}. Rainbow Teaming emphasizes diversity~\cite{rainbow}, WildTeaming derives strategies from observed jailbreaks~\cite{wildteaming}, and AutoDAN-Turbo reuses discovered strategies~\cite{autodanturbo}. ActorAttack studies multi-turn interactions~\cite{actorattack}. These lines primarily determine which candidates or interaction sequences to try.

RedAgent uses contextual memory~\cite{redagent}. ALI-Agent generates and refines alignment test scenarios~\cite{aliagent}. EMJO coordinates an attacker, analyzer, and judge through an experience bank~\cite{emjo}. AutoRedTeamer integrates additional attack techniques~\cite{autoredteamer}, and AJAR orchestrates procedures exposed through a common interface~\cite{ajar}. These methods use experience and execution feedback to adapt the testing process.

AgenticRed uses evolutionary selection and generational knowledge to discover reusable testing systems~\cite{agenticred}. EvoSynth searches executable attack programs and uses execution feedback to repair them~\cite{evosynth}. These approaches make the testing system itself an object of optimization. L2C examines a decision inside an ongoing case. Its learned controller uses graph-versioned records to decide which organizational fields to edit and whether another round of editing is useful. The execution engine resolves the edited bindings so that a change in information flow can be checked against the next pass's inputs.

\subsection{Workflow Design and Orchestration}
AutoGen represents agent interactions as programmable conversations~\cite{autogen}, MetaGPT specifies collaboration through operating procedures~\cite{metagpt}, and MacNet studies directed acyclic topologies~\cite{macnet}. ADAS searches agentic systems~\cite{adas}, and AFlow searches executable workflows~\cite{aflow}. MaAS learns query-dependent architectures and resource allocations~\cite{maas}. AOrchestra learns to configure and delegate to sub-agents~\cite{aorchestra}.

GRAFT revises selected regions of a globally optimized workflow using execution-quality signals and consistency checks~\cite{graft}. InFlowOp combines construction with runtime correction using a label-free cost that reflects task demands, agent competence, and execution expenditure~\cite{inflowop}. L2C shares their focus on selective runtime adaptation. In red teaming, the external verdict evaluates the target's response, so the controller pairs that verdict with the node inputs, outputs, and dependencies that produced the submitted candidate. This evidence links an edit to a concrete change in subsequent execution.

\subsection{Feedback and Evaluation}
Self-Refine revises outputs using generated feedback~\cite{selfrefine}, and Reflexion carries linguistic reflections into later attempts~\cite{reflexion}. In a multi-agent workflow, explicit input records also show where that feedback was delivered. HarmBench~\cite{harmbench} and JailbreakBench~\cite{jailbreakbench} standardize behavior-level evaluation. StrongREJECT distinguishes apparent compliance from substantive fulfillment~\cite{strongreject}. L2C records workflow evidence alongside verdicts. A reduction in target requests can coexist with more tokens spent on planning or exchanges, which motivates counting coordination, execution, and assessment in the same measure.

\section{Method}
\label{sec:method}

\begin{figure*}[!t]
\centering
\includegraphics[width=\textwidth]{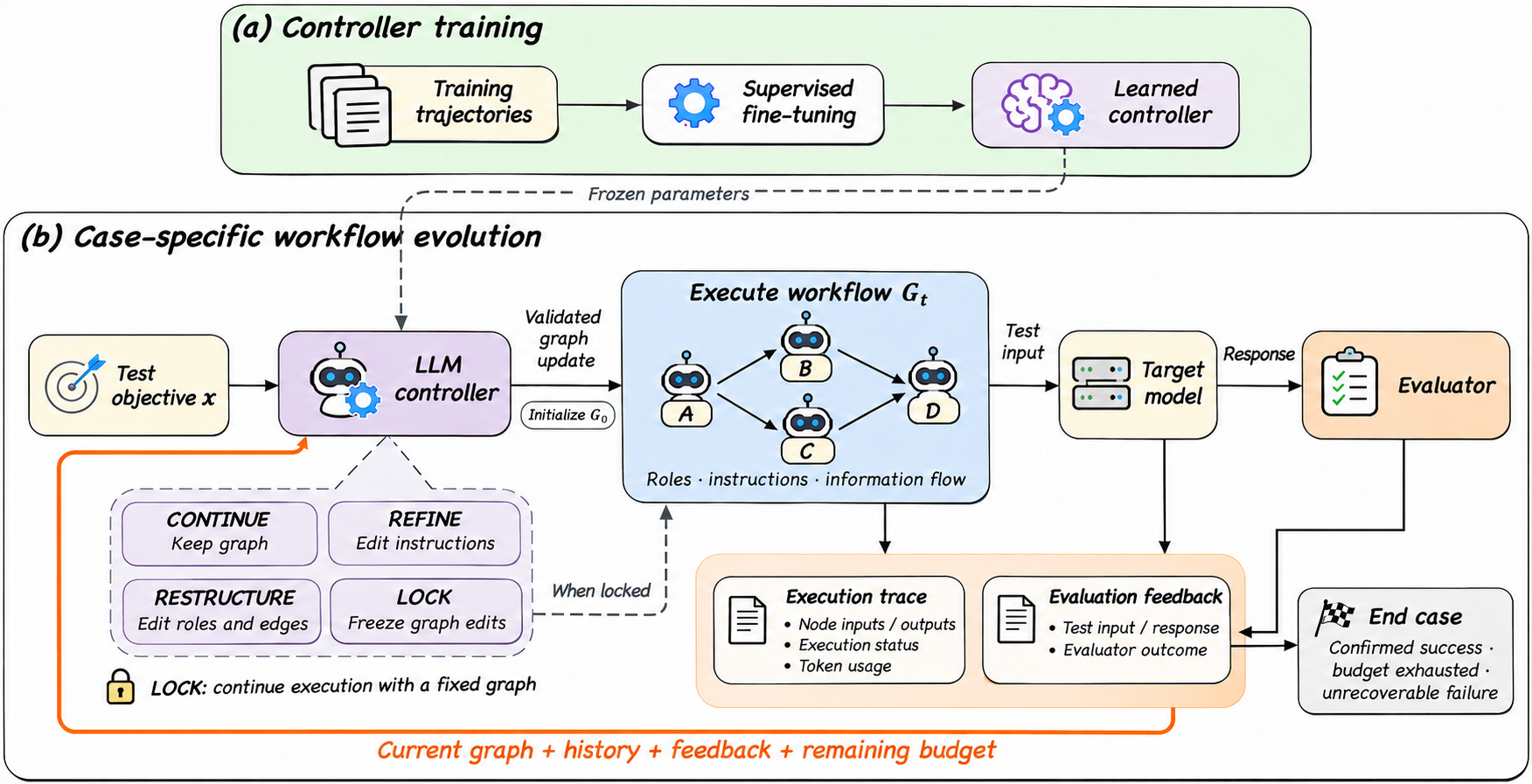}
\caption{Overview of L2C. (a) Execution trajectories supervise controller training. (b) During evaluation, the frozen controller uses case-local traces, verdicts, and budget to select changes before the next pass. LOCK ends editing while execution continues.}
\label{fig:overview}
\end{figure*}

\subsection{Problem Setting}
A case specifies a fixed behavioral objective $x$, a target model $M$, and a response evaluator $J$~\cite{harmbench,jailbreakbench}. L2C accesses the target through its input--output interface. The controller organizes the workflow, and execution agents realize its nodes using a designated execution backbone. The evaluator scores target responses against $x$. Each episode has an online token allowance $B$. Figure~\ref{fig:overview} shows how the workflow is revised using case-local feedback with the controller parameters frozen.

An episode contains successive workflow passes. A pass is the internal execution of the current graph and supplies a candidate for evaluation. The target parameters, behavioral objective, and evaluator remain fixed throughout the episode. We assess the success--cost trade-off through the probability of evaluator-confirmed success and the expected number of online tokens consumed under $B$. Both measures account for episodes that end without a successful response.

\subsection{Executable Collaboration Graph}
\label{sec:graph}
The workflow at decision step $t$ is
\begin{equation}
G_t=(V_t,E_t,\rho_t,o_t).
\end{equation}
Each node $v\in V_t$ has a stable identifier, responsibility $r_v$, instruction $p_v$, named input slots $\mathcal I_v$, and an output contract $\mathcal O_v$. The responsibility defines the node's function, while the instruction describes how the node carries it out. The edge set $E_t$ orders execution. An input binding $\rho_t(v,k)$ names the artifact supplied to slot $k\in\mathcal I_v$, and $o_t$ identifies the artifact submitted to the target.

The executor traverses the graph in topological order. A binding can reference an earlier node's output in the current pass or a versioned external record, such as the objective or a prior pass's feedback. Current-pass bindings require the producer to precede the consumer in $E_t$. These two reference types allow case history to persist across acyclic passes. Each pass invokes the nodes with their instructions and resolved inputs.

The controller initializes $G_0$ with a valid submission reference and empty execution history. Subsequent edits take effect at pass boundaries. A newly added node receives a new identifier, and a retained node keeps its identifier so that successive invocations can be compared. Each pass uses one graph version for all of its node invocations.

\begin{figure*}[t]
\centering
\includegraphics[width=\textwidth]{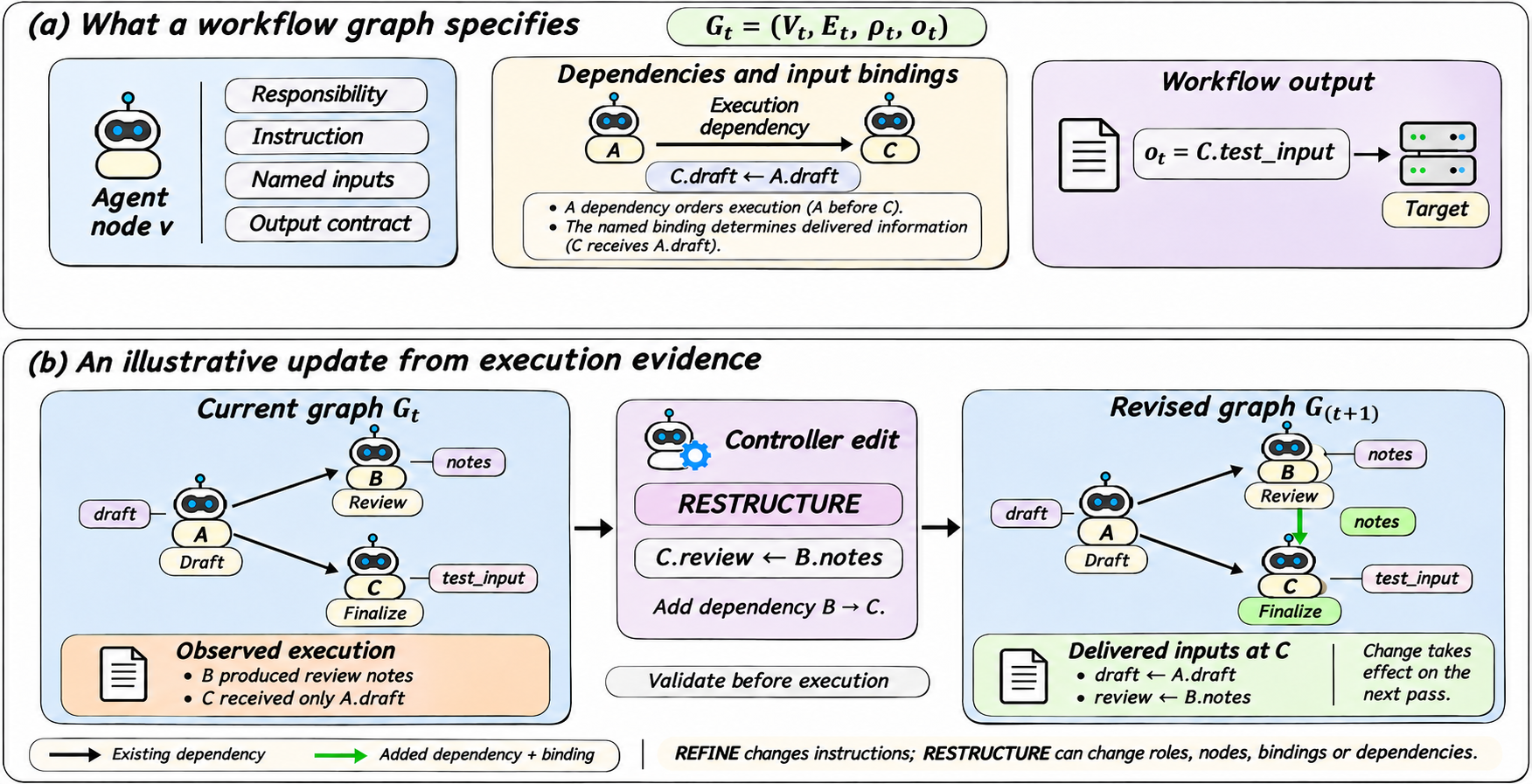}
\caption{Executable graph and illustrative update. (a) Dependencies order execution, and bindings specify delivered information. (b) C's recorded inputs lack B's review notes. The controller adds a B~$\rightarrow$~C dependency and a binding that delivers the notes on the next pass.}
\label{fig:graph-evolution}
\end{figure*}

A validator checks identifier uniqueness, references, interfaces, acyclicity, and the submission reference on a copy of the proposed graph. The update commits only when all checks pass. A rejected patch returns the affected fields and failure type. The trace preserves the graph version, node identifier, resolved input sources, content delivered, artifacts produced, execution status, and token usage. Recording actual delivery lets the controller distinguish the declared binding from the data available to the node. Figure~\ref{fig:graph-evolution} illustrates how a dependency and a binding serve different purposes in an edit.

\subsection{Execution Feedback}
\label{sec:feedback}
The controller receives three kinds of feedback. Outcome records link the submitted candidate, target response, and evaluator verdict. Execution records describe what each node received and produced, including node failures and validator messages. Resource records give the tokens used and remaining. The verdict measures target response quality under the benchmark criterion, while invocation records show the workflow evidence relevant to an edit. An executor refusal is stored with its node invocation, and a target refusal is stored with the target interaction. Serialized records and their summaries retain source identifiers and graph versions, which the controller can cite when proposing a change.

The controller's diagnosis is recorded as an interpretation of these observations. For example, execution records can establish that a bound artifact was absent. The effect of changing its delivery on the final outcome requires another execution. Keeping the verdict, trace, and diagnosis associated with the correct pass lets the controller distinguish its observations from proposed explanations.

\subsection{Controller Decisions}
\label{sec:controller}
The controller observes $s_t=(x,G_t,H_t,F_t,b_t,\ell_t)$, with case history $H_t$, latest feedback $F_t$, remaining tokens $b_t$, and editing-lock indicator $\ell_t$. It selects $a_t=(u_t,\Delta_t,\eta_t)$ under policy $\pi_\theta(\cdot\mid s_t)$. The action specifies its type $u_t$, patch $\Delta_t$, and supporting record references $\eta_t$. The accepted patch controls execution, and the references identify the evidence for the decision.

\begin{itemize}
\item \textsc{Continue} executes the current graph and permits editing decisions after the pass.
\item \textsc{Refine} changes node instructions while preserving identifiers, responsibilities, interfaces, dependencies, bindings, and the submission reference.
\item \textsc{Restructure} changes node membership, responsibilities, interfaces, dependencies, bindings, or the submission reference. An edit can include related instruction changes.
\item \textsc{Lock} preserves the graph and bypasses further controller calls. Execution continues with updated artifacts through the fixed bindings.
\end{itemize}

The patch is empty for \textsc{Continue} and \textsc{Lock}. A structural edit changes a field used during execution. A new candidate produced by the same graph is a different execution outcome. This distinction identifies graph adaptation independently of changes in candidate text and supports the fixed-organization comparison in Section~\ref{sec:ablations}.

Let $\widetilde G_t$ be the graph retained or produced by a validated action, and let $\tau_t$ record the resulting pass, target interaction, and evaluation. An accepted action gives
\begin{equation}
\begin{aligned}
G_{t+1}&=\widetilde G_t, \quad
H_{t+1}=H_t\oplus(a_t,\widetilde G_t,\tau_t),\\
F_{t+1}&=\operatorname{Feedback}(\tau_t), \quad
b_{t+1}=b_t-c_t,\\
\ell_{t+1}&=\ell_t\lor\mathbf{1}[u_t=\textsc{Lock}],
\end{aligned}
\end{equation}
where $\oplus$ appends to the history and $c_t$ counts decision, execution, and online evaluation tokens. A rejected proposal adds validation feedback and consumes its incurred cost without advancing the workflow pass or submitting a candidate.

When locked, the graph executes with an implicit \textsc{Continue} action. Its bindings continue resolving fresh case records, so execution can react to new feedback even after the controller stops editing. An unlocked \textsc{Continue} leaves open a future editing decision. The episode ends on confirmed success, budget exhaustion, or unrecoverable failure. The two stopping decisions govern different parts of the process. Locking ends graph adaptation, and episode termination ends evaluation.

\subsection{Supervised Learning of Coordination Decisions}
\label{sec:training}
The training set combines interface demonstrations $\mathcal D_{\mathrm{int}}$ and execution-grounded decisions $\mathcal D_{\mathrm{exec}}$. Interface demonstrations pair a graph with a valid action to teach the schema and edit constraints. They cover graph retention, instruction edits, structural changes, and locking. Execution-grounded decisions pair a pre-action state with an action selected using subsequent success and cost. This formulation builds on supervised learning for orchestration~\cite{aorchestra}.

For a selected state in a training episode, we compare admissible actions from the same graph, history, feedback, and remaining allowance. The branches use the same models, evaluator, and continuation policy. A locking branch uses the frozen graph for its remaining execution. Other actions allow further decisions under the continuation policy. For action $a$ in state $s$, the $K$ continuations estimate
\begin{equation}
\widehat p(s,a)=\frac{1}{K}\sum_{k=1}^{K}y_k(s,a),\qquad
\widehat c(s,a)=\frac{1}{K}\sum_{k=1}^{K}c_k(s,a),
\end{equation}
where $y_k$ denotes within-budget success and $c_k$ counts additional online tokens, including later controller decisions. Relative to the continue action $a_0$, a candidate improves either success with controlled cost,
\begin{equation}
\begin{aligned}
\widehat p(s,a)&>\widehat p(s,a_0)+\epsilon_p,\\
\widehat c(s,a)&\leq(1+\epsilon_c)\widehat c(s,a_0),
\end{aligned}
\label{eq:success-selection}
\end{equation}
or cost with controlled success,
\begin{equation}
\begin{aligned}
\widehat p(s,a)&\geq\widehat p(s,a_0)-\epsilon_p,\\
\widehat c(s,a)&<(1-\epsilon_c)\widehat c(s,a_0).
\end{aligned}
\label{eq:cost-selection}
\end{equation}
Validation fixes $\epsilon_p$ and $\epsilon_c$. Eligible actions are ranked by success, with cost deciding comparisons within the success tolerance. The collection rule retains stable labels and leaves unresolved or all-failed states unlabeled. For instance, a branch that terminates early and spends fewer tokens provides no positive coordination label when all branches fail. Continuing is labeled only when the comparisons support retaining the workflow. This selection keeps structurally valid demonstrations and empirically supported decisions distinct. Appendix~\ref{app:reproducibility} specifies label selection and training settings.

For $\mathcal D=\mathcal D_{\mathrm{int}}\cup\mathcal D_{\mathrm{exec}}$, supervised fine-tuning minimizes the action-token loss
\begin{equation}
\mathcal L_{\mathrm{coord}}(\theta)
=-\mathbb E_{(s,a)\sim\mathcal D}
\sum_{j=1}^{|a|}\log\pi_\theta(a_j\mid s,a_{<j}).
\end{equation}
The loss covers the action type, patch, and cited evidence, with the pre-action state as conditioning context. The branch outcomes select the action label offline. Training and validation cases are assigned before trajectory collection to keep each case's branches in the same partition. At test time the controller parameters and decision settings are frozen. Each case starts with its own history and the controller selects the next action directly from the current state.

\subsection{Online Resource Accounting}
\label{sec:accounting}
For each episode $i$, the set $\mathcal C_i$ includes every online call by the controller, executor, target, and evaluator. The token cost is
\begin{equation}
T_i=\sum_{c\in\mathcal C_i}\left(t_c^{\mathrm{in}}+t_c^{\mathrm{out}}\right).
\label{eq:tokens}
\end{equation}
Inputs include instructions, graph descriptions, bound artifacts, and reused history each time they are processed. Output counts include backend-reported reasoning tokens according to the backend's usage records. Initialization, edits, feedback summarization, repairs, and failed calls enter the same sum. We report target requests separately because each request can require different amounts of coordination and execution.

Online token volumes are compared within an execution configuration. Offline trajectory collection, system search, and fine-tuning are reported separately. An evaluator call that informs a subsequent decision belongs to the online cost, while independent re-scoring of saved responses belongs to final assessment. This boundary includes all work spent on coordinating a test even when it occurs outside a node invocation.

\section{Experiments}
\label{sec:experiments}
% Numerical cells imported without alteration from the author-supplied tables dated 2026-11-17.
% Reconcile their run manifests with the expanded evaluation protocol before submission.
\begin{table*}[t]
\centering
\caption{HarmBench results across execution backbones and target models with an 8M online-token allowance. ASR is the attack success rate; MTok and $Q$ are mean total online tokens (millions) and target requests. Bold marks the best value within each backbone; underlining highlights L2C's costs. GRAFT-HB is the task adaptation in Section~\ref{sec:experimental-setup}.}
\label{tab:main}
\begin{tabular*}{\textwidth}{@{\extracolsep{\fill}}lrrrrrr@{}}
\toprule
& \multicolumn{3}{c}{Llama-3.1-70B-Instruct} & \multicolumn{3}{c}{GPT-5.4} \\
\cmidrule(lr){2-4}\cmidrule(lr){5-7}
Method & ASR (\%) $\uparrow$ & MTok $\downarrow$ & $Q$ $\downarrow$ & ASR (\%) $\uparrow$ & MTok $\downarrow$ & $Q$ $\downarrow$ \\
\midrule
\multicolumn{7}{l}{\textit{Execution backbone: Mixtral-8x7B-Instruct-v0.1}} \\
PAIR~\cite{pair} & 68.5 & \textbf{2.45} & 18.4 & 59.0 & \textbf{2.85} & 21.6 \\
AutoDAN-Turbo~\cite{autodanturbo} & 76.2 & 4.10 & 15.6 & 68.5 & 4.65 & 18.2 \\
AutoRedTeamer~\cite{autoredteamer} & 79.0 & 3.85 & 13.8 & 71.0 & 4.30 & 15.9 \\
AgenticRed~\cite{agenticred} & 82.5 & 5.60 & 11.5 & 75.5 & 6.20 & 13.2 \\
EvoSynth~\cite{evosynth} & 83.1 & 5.20 & 10.8 & 76.5 & 5.85 & 12.4 \\
GRAFT-HB~\cite{graft} & 82.0 & 5.45 & 11.2 & 75.2 & 6.05 & 12.9 \\
L2C & \textbf{84.5} & \underline{3.35} & \underline{\textbf{7.6}} & \textbf{78.0} & \underline{3.75} & \underline{\textbf{8.8}} \\
\midrule
\multicolumn{7}{l}{\textit{Execution backbone: Qwen3.5-4B}} \\
PAIR~\cite{pair} & 61.5 & \textbf{2.65} & 19.8 & 51.5 & \textbf{3.05} & 23.1 \\
AutoDAN-Turbo~\cite{autodanturbo} & 69.5 & 4.45 & 17.1 & 60.5 & 4.95 & 19.5 \\
AutoRedTeamer~\cite{autoredteamer} & 72.5 & 4.15 & 15.2 & 63.5 & 4.65 & 17.4 \\
AgenticRed~\cite{agenticred} & 75.0 & 6.05 & 13.1 & 66.5 & 6.65 & 14.8 \\
EvoSynth~\cite{evosynth} & 75.8 & 5.65 & 12.5 & 67.0 & 6.30 & 14.2 \\
GRAFT-HB~\cite{graft} & 74.6 & 5.90 & 12.8 & 65.8 & 6.50 & 14.5 \\
L2C & \textbf{79.5} & \underline{3.55} & \underline{\textbf{8.4}} & \textbf{71.5} & \underline{3.95} & \underline{\textbf{9.6}} \\
\midrule
\multicolumn{7}{l}{\textit{Execution backbone: Gemini 3.5 Flash-Lite}} \\
PAIR~\cite{pair} & 66.8 & \textbf{2.15} & 17.9 & 57.2 & \textbf{2.50} & 20.8 \\
AutoDAN-Turbo~\cite{autodanturbo} & 74.5 & 3.60 & 15.0 & 66.4 & 4.05 & 17.5 \\
AutoRedTeamer~\cite{autoredteamer} & 77.2 & 3.35 & 13.2 & 69.0 & 3.75 & 15.2 \\
AgenticRed~\cite{agenticred} & 80.6 & 4.85 & 11.0 & 73.4 & 5.40 & 12.6 \\
EvoSynth~\cite{evosynth} & 81.2 & 4.50 & 10.4 & 74.2 & 5.05 & 11.9 \\
GRAFT-HB~\cite{graft} & 80.1 & 4.70 & 10.7 & 73.0 & 5.25 & 12.3 \\
L2C & \textbf{83.2} & \underline{2.90} & \underline{\textbf{7.2}} & \textbf{76.8} & \underline{3.25} & \underline{\textbf{8.3}} \\
\bottomrule
\end{tabular*}
\end{table*}

\subsection{Experimental Setup}
\label{sec:experimental-setup}
\textbf{Data.} WildJailbreak objectives~\cite{wildteaming} are grouped before trajectory collection so that a request and its derived variants share a split. The protocol allocates 512 objective groups to training and 128 to validation. Training supplies demonstrations and coordination labels, and validation selects the controller checkpoint and comparison tolerances. HarmBench Standard is the primary test benchmark~\cite{harmbench}. JailbreakBench provides a transfer test, including 60 behaviors disjoint from HarmBench~\cite{jailbreakbench}. Objectives overlapping either benchmark are excluded from controller development. Appendix~\ref{app:reproducibility} describes grouping and supervision records.

\textbf{Models.} The controller is Mixtral-8x22B-Instruct-v0.1~\cite{mixtral22} fine-tuned with LoRA~\cite{lora}. Trajectory collection and validation use the Mixtral executor and Llama target. Evaluation pairs Mixtral-8x7B-Instruct-v0.1~\cite{mixtral7}, Qwen3.5-4B~\cite{qwen35}, and Gemini 3.5 Flash-Lite~\cite{gemini35lite} with Llama-3.1-70B-Instruct~\cite{llama31} and GPT-5.4~\cite{gpt54}. The six configurations share the selected controller checkpoint, action interface, and hyperparameters.

\textbf{Baselines.} We compare with PAIR~\cite{pair}, AutoDAN-Turbo~\cite{autodanturbo}, AutoRedTeamer~\cite{autoredteamer}, AgenticRed~\cite{agenticred}, and EvoSynth~\cite{evosynth}. These cover prompt refinement, strategy reuse, attack integration, system evolution, and executable-program evolution. GRAFT-HB adapts GRAFT's workflow-region replacement and consistency checks~\cite{graft} to the shared target interface, using the same outcome and token measures. The suffix HB identifies this task adaptation of the original method.

Within each configuration, methods share executor and target models, evaluator, development pools, and online budget. Separate meta-design roles, where used, run on the unadapted controller backbone. Libraries and searched workflows are frozen after development. Each method uses its native within-episode history and initialization, with development expenditure recorded separately.

\textbf{Evaluation.} The HarmBench protocol specifies 159 Standard behaviors, three independent episodes per behavior, and $B=8\times10^6$ online tokens per episode for the main comparison. Every submitted candidate starts a fresh target conversation under a fixed target system prompt and chat template. Internal history persists within the episode and resets across cases and replicates.

The frozen HarmBench-Llama-2-13b-cls evaluator uses the official Standard template and deterministic decoding~\cite{harmbench}. It receives the behavior and target response with method identity and workflow records concealed. A positive verdict establishes success within the budget. Executor refusals, target refusals, invalid proposals, and assessment errors are recorded separately. Appendix~\ref{app:reproducibility} details inference settings and call accounting.

\textbf{Metrics.} For $N$ behaviors and $R$ replicates, let $y_{ir}$ denote episode success, $T_{ir}$ the token count in Eq.~\eqref{eq:tokens}, and $Q_{ir}$ the number of dispatched target requests. At the specified budget we report
\begin{equation}
\begin{aligned}
\ASR &= \frac{100}{NR}\sum_{i,r}y_{ir},\\
\mathrm{MTok} &= \frac{1}{10^6 NR}\sum_{i,r}T_{ir},\qquad
\overline Q = \frac{1}{NR}\sum_{i,r}Q_{ir}.
\end{aligned}
\label{eq:empirical-metrics}
\end{equation}
Failed episodes enter all denominators and retain their incurred costs. $Q$ counts every dispatched target request, including retries. The tables report point estimates. Appendix~\ref{app:additional-results} describes the behavior-level resampling protocol.

\subsection{Effectiveness and Token Efficiency}
\label{sec:main-results}
\begin{table*}[t]
\centering
\caption{Primary-configuration costs in millions of tokens. Component means, P50, and P90 use 8M episodes; $C_{75}$ uses the full budget grid. Component means sum to Total. NR denotes that 75\% ASR is not reached.}
\label{tab:cost-components}
\begin{tabular*}{\textwidth}{@{\extracolsep{\fill}}lrrrrrrrr@{}}
\toprule
Method & Controller & Executor & Target & Evaluator & Total & P50 & P90 & $C_{75}$ \\
\midrule
PAIR & 0.00 & 2.15 & 0.18 & 0.12 & 2.45 & 1.95 & 5.40 & NR \\
AutoDAN-Turbo & 0.00 & 3.82 & 0.17 & 0.11 & 4.10 & 3.10 & 7.85 & 4.10 \\
AutoRedTeamer & 0.45 & 3.15 & 0.15 & 0.10 & 3.85 & 2.85 & 7.60 & 3.30 \\
AgenticRed & 1.65 & 3.72 & 0.14 & 0.09 & 5.60 & 4.40 & 8.00 & 3.65 \\
EvoSynth & 1.45 & 3.53 & 0.13 & 0.09 & 5.20 & 4.05 & 8.00 & 3.40 \\
GRAFT-HB & 1.15 & 4.07 & 0.14 & 0.09 & 5.45 & 4.25 & 8.00 & 3.75 \\
L2C & 0.65 & 2.52 & 0.11 & 0.07 & \underline{3.35} & 2.20 & 6.85 & \underline{2.40} \\
\bottomrule
\end{tabular*}
\end{table*}

Table~\ref{tab:main} reports HarmBench ASR, total online tokens, and target requests at the common 8M allowance. The six configurations include the development setting, an executor change, a target change, and both changes together. Comparisons use each configuration's reported ASR, MTok, and $Q$. The controller can revise the graph within each episode with its parameters held fixed.

\textbf{Matched-success cost.} We compare token expenditure at a common success level using independent runs at allowances $\mathcal B=\{2,3,4,5,6,8\}\times10^6$ with the Mixtral executor and Llama target. Each budget is set at initialization, since it can influence subsequent decisions. An operating point consists of the achieved ASR and actual mean expenditure for that allowance. For $\alpha=75\%$, matched-success cost is
\begin{equation}
C_\alpha=\min_{B\in\mathcal B:\,\ASR(B)\geq\alpha}\overline T(B),
\label{eq:matched-cost}
\end{equation}
where $\overline T(B)$ is mean online token use at budget $B$. We report NR when no operating point reaches the threshold. The minimum uses the measured operating points on the shared grid.

\textbf{Cost components.} Table~\ref{tab:cost-components} reports $C_{75}$ and decomposes the 8M token total into controller, executor, target, and evaluator components. Costs are assigned by function, with zero controller tokens for methods that have no separate controller. The means cover the same episodes and sum to total MTok. P50 and P90 summarize the episode-level distribution. Appendix~\ref{app:reproducibility} reports offline inference and fine-tuning resources separately.
% Results and analysis.

\subsection{Key Ablations}
\label{sec:ablations}
The ablations use Mixtral-8x7B-Instruct-v0.1 and Llama-3.1-70B-Instruct with the main comparison's cases, replicates, and evaluator. Variants share a case-conditioned initial graph for each behavior and replicate, and every variant includes its generation cost. Action restrictions are enforced by the interface, keeping the comparison focused on the available decisions after initialization.

\emph{Fixed organization} permits instruction refinement while masking structural edits. \emph{Outcome-only feedback} preserves the objective, graph, target response, verdict, and budget, and removes node traces and derived summaries during training and inference. \emph{Untrained controller} uses the base model after initialization. \emph{Without LOCK} disables persistent locking while retaining continued execution, instruction refinement, and structural edits.

Table~\ref{tab:ablation} compares the four interventions. In addition to success and resource use, it reports rejected proposals as a proportion of submitted proposals. This provides a measure of interface validity alongside the outcome metrics. Appendix~\ref{app:additional-results} reports the frozen workflow, unlinked trace, and interface-only SFT variants.
\begin{table}[t]
\centering
\caption{Core ablations with the Mixtral executor and Llama target. Metrics use 8M except $C_{75}$, which uses the budget grid. ASR and invalid proposals are percentages. MTok and $C_{75}$ are in millions of tokens. Bold marks the best value.}
\label{tab:ablation}
\setlength{\tabcolsep}{3pt}
\begin{tabular*}{\columnwidth}{@{\extracolsep{\fill}}lrrrrr@{}}
\toprule
Variant & ASR & MTok & $Q$ & $C_{75}$ & Invalid \\
\midrule
L2C & \textbf{84.5} & \underline{\textbf{3.35}} & \underline{\textbf{7.6}} & \underline{\textbf{2.40}} & \textbf{1.5} \\
Fixed organization & 80.2 & 3.90 & 9.8 & 3.35 & 2.1 \\
Outcome-only feedback & 81.5 & 4.40 & 10.5 & 3.65 & 3.4 \\
Untrained controller & 75.0 & 4.85 & 13.2 & 4.85 & 14.8 \\
Without LOCK & 84.3 & 4.25 & 7.8 & 3.15 & 1.9 \\
\bottomrule
\end{tabular*}
\end{table}

% Results and analysis.

\subsection{Dataset Transfer}
\label{sec:transfer}
The JailbreakBench evaluation keeps the Mixtral--Llama setup, classifier, and 8M allowance. Methods retain their development artifacts and run on the new behavior set with the same scoring protocol. Table~\ref{tab:transfer} compares the full 100-behavior set and the 60-behavior subset disjoint from HarmBench. The latter measures transfer after removing overlap with the primary benchmark.
\begin{table}[!t]
\centering
\caption{JailbreakBench transfer (Mixtral--Llama, three replicates, 8M). ASR is in percent.}
\label{tab:transfer}
\setlength{\tabcolsep}{3pt}
\begin{tabular*}{\columnwidth}{@{\extracolsep{\fill}}lrrrrrr@{}}
\toprule
& \multicolumn{3}{c}{Full ($N=100$)} & \multicolumn{3}{c}{Disjoint ($N=60$)} \\
\cmidrule(lr){2-4}\cmidrule(lr){5-7}
Method & ASR & MTok & $Q$ & ASR & MTok & $Q$ \\
\midrule
PAIR & 71.0 & \textbf{2.30} & 17.2 & 68.3 & \textbf{2.42} & 18.0 \\
AutoDAN-Turbo & 78.5 & 3.95 & 14.8 & 75.0 & 4.15 & 15.6 \\
AutoRedTeamer & 81.0 & 3.70 & 13.1 & 77.8 & 3.88 & 13.8 \\
AgenticRed & 84.0 & 5.40 & 10.9 & 81.2 & 5.65 & 11.5 \\
EvoSynth & 84.8 & 5.05 & 10.2 & 81.7 & 5.30 & 10.8 \\
GRAFT-HB & 83.5 & 5.25 & 10.6 & 80.5 & 5.50 & 11.2 \\
L2C & \textbf{86.3} & \underline{3.20} & \underline{\textbf{7.1}} & \textbf{84.5} & \underline{3.32} & \underline{\textbf{7.5}} \\
\bottomrule
\end{tabular*}
\end{table}

% Results and analysis.

\subsection{Evaluator Reliability}
\label{sec:judge-study}
\begin{table}[!t]
\centering
\caption{Independent StrongREJECT scores on saved 8M episodes. Each episode contributes its highest response score. Higher is better.}
\label{tab:independent-scores}
\setlength{\tabcolsep}{4pt}
\begin{tabular*}{\columnwidth}{@{\extracolsep{\fill}}lrrrr@{}}
\toprule
 & \multicolumn{2}{c}{Mixtral executor} & \multicolumn{2}{c}{Qwen executor} \\
\cmidrule(lr){2-3}\cmidrule(lr){4-5}
Method & Llama & GPT & Llama & GPT \\
\midrule
PAIR & 0.512 & 0.425 & 0.448 & 0.365 \\
AutoDAN-Turbo & 0.594 & 0.510 & 0.525 & 0.442 \\
AutoRedTeamer & 0.628 & 0.542 & 0.558 & 0.475 \\
AgenticRed & 0.665 & 0.585 & 0.586 & 0.504 \\
EvoSynth & 0.674 & 0.596 & 0.595 & 0.512 \\
GRAFT-HB & 0.660 & 0.580 & 0.582 & 0.498 \\
L2C & \textbf{0.698} & \textbf{0.624} & \textbf{0.635} & \textbf{0.558} \\
\bottomrule
\end{tabular*}
\end{table}

The online verdict determines episode success and therefore influences when each episode stops. We use the released StrongREJECT evaluator~\cite{strongreject} to independently re-score saved target responses from the four Mixtral/Qwen configurations. The scorer is held out from controller training, baseline development, and online stopping. Each episode contributes its maximum response score, or zero if it produces no target response. Table~\ref{tab:independent-scores} averages these continuous scores over behaviors and replicates. The comparison concerns the responses collected under the original protocol.
A complementary blinded audit measures agreement with human judgments on individual responses. Appendix~\ref{app:additional-results} gives the audit protocol and reports agreement alongside false-positive and false-negative rates. The run records distinguish executor refusals, target refusals, invalid proposals, and budget exhaustion to separate construction, response, and evaluation failures.

\section{Conclusion}
Learning to Coordinate (L2C) uses execution feedback to adapt multi-agent red-teaming workflows for individual cases. By linking each graph version to what agents receive and produce, L2C makes it possible to trace how a change in collaboration affects the next execution. Its controller also learns when to stop editing and continue with the current graph, so adaptation itself is part of the resource decision.

Across six HarmBench configurations, L2C improves attack success over EvoSynth by 1.4--4.5 percentage points while using 35.6--37.3\% fewer online tokens on average. In the primary configuration, reaching the 75\% ASR threshold costs 2.40 million tokens per case, compared with 3.30 million for the lowest-cost baseline that reaches the same threshold. The ablations suggest complementary roles for the design choices. Structural edits and node-level feedback improve observed success, while locking primarily reduces token use.

Future work can examine whether these token savings translate into lower latency and monetary cost across inference backends. Evaluation with additional independent judges can also test how sensitive the success--cost trade-off is to the scoring criterion.

\bibliographystyle{IEEEtran}
\bibliography{references}
\appendices
\section{Experimental Configuration and Reproducibility}
\label{app:reproducibility}
\subsection{Objective Grouping and Data Records}
Objective normalization applies Unicode NFKC normalization, case folding, and whitespace normalization. Exact matches and pairs with token 5-gram Jaccard similarity of at least 0.8 form connected groups; dataset-provided derivation identifiers additionally join variants of the same underlying objective. Cross-dataset near matches are manually checked before splitting. Filtering uses objective content without model responses or test scores. The split manifest records original identifiers, group membership, excluded overlaps, and split assignment. Training and validation groups are sampled with seed 2026. The 512/128 allocation describes the objective pools, not the number of accepted SFT examples. Accepted examples depend on trajectory validity and decision-label stability.

\subsection{Controller Supervision and Optimization}
The collection policy is a frozen copy of the base controller. Initialization and interface demonstrations use the graph schema in Section~\ref{sec:graph}. For each training objective, the first completed, nonterminal workflow pass supplies a candidate decision state when available. One validated instruction-edit proposal and one validated structural-edit proposal are compared with CONTINUE and LOCK. Invalid candidates are recorded and excluded from branch execution. Each valid branch receives $K=5$ continuations under the same frozen continuation policy, execution models, evaluator, and remaining allowance. Candidate proposals and their preparation costs are included in offline collection expenditure.

Decision-label stability is checked by recomputing the selected action after leaving out each continuation replicate. A label is retained only if all five checks agree with the full-sample selection. The validation search considers $\epsilon_p\in\{0,0.1\}$ and $\epsilon_c\in\{0.05,0.1\}$ in Eqs.~\eqref{eq:success-selection}--\eqref{eq:cost-selection}. For eligible candidates, success outside the tolerance has priority over cost; unresolved ties remain unlabeled. A CONTINUE target additionally requires observed success and support for retaining the graph under the same comparisons. States with no reliable label remain in the collection record. Label counts, action frequencies, and the selected tolerance pair are reported as collection outcomes in Table~\ref{tab:training-record}.

LoRA adapters apply to attention projections and expert feed-forward linear layers; router, embeddings, and original weights remain frozen. The specified configuration uses rank 32, scaling factor 64, adapter dropout 0.05, bfloat16 arithmetic, and gradient checkpointing. Optimization uses AdamW, a learning rate of $2\times10^{-5}$, effective batch size 16, two epochs, 3\% warm-up, cosine decay, weight decay 0.01, and gradient clipping at norm 1.0. Checkpoints at epoch boundaries are compared on validation ASR, with lower mean tokens breaking ties. The training seed is 2026. Interface and execution-selected examples receive equal sampling probability at the source level. Supervision applies only to action tokens.

\begin{table}[t]
\centering
\caption{Observed supervision and training records. Configured objective pools are specified in the text; counts below concern collected and retained examples.}
\label{tab:training-record}
\begin{tabularx}{\columnwidth}{@{}Xr@{}}
\toprule
Item & Observed value \\
\midrule
Interface demonstrations & 300 \\
Candidate decision states & 428 \\
Valid continuation branches & 1,540 \\
Retained execution-selected decisions & 296 \\
CONTINUE / REFINE labels & 112 / 84 \\
RESTRUCTURE / LOCK labels & 58 / 42 \\
Selected $(\epsilon_p,\epsilon_c)$ & (0.0, 0.05) \\
Training action tokens & 0.42M \\
SFT GPU-hours & 18.5 \\
\bottomrule
\end{tabularx}
\end{table}

\subsection{Inference Configuration and Budget Admission}
Controller and executor sampling use temperature 0.7 and top-$p$ 0.9, fixed across methods for each role. The Llama target uses greedy decoding. GPT target calls use a fixed provider model version and provider-supported decoding configuration throughout the comparison. Returned model identifiers, request settings, tokenizer revisions, and inference-library versions are stored in the run manifest. Validation episodes use seeds 101, 102, and 103; test episodes use 17, 29, and 43 where the backend supports explicit seeds. Independent API episodes are retained even when the provider cannot guarantee deterministic seed replay.

Node fields and outputs are not shortened to enforce an arbitrary schema-length cap. Inputs must fit the backend's context window; any history selection or summarization is logged and its inference cost is charged. Before dispatch, the budget manager accounts for input tokens and limits permitted completion expenditure to the remaining allowance. Candidate generation additionally reserves enough budget to assess the resulting target response. Unassessed responses do not establish within-budget success. Backend-reported reasoning-token usage is included exactly once, respecting whether it is already contained in the output total.

Transport errors permit at most two retries with the same request and episode identity. Every dispatched target request counts toward $Q$, and all reported usage is charged. Failures with unavailable usage are flagged; an affected comparison block is repeated under the same case and seed schedule after service recovery rather than assigned zero cost. Invalid judge output is retried once with identical content; persistent failure terminates the episode as an evaluation error. Semantic refusals and exhausted budgets do not trigger replacement episodes. Native context-window violations and unrecoverable execution errors remain recorded failures.

\subsection{Artifacts and Offline Resource Records}
Run records contain behavior and replicate identifiers, graph versions, accepted and rejected patches, resolved inputs, outputs, response verdicts, stopping reasons, and per-call usage. Frozen checkpoints, adapters, baseline libraries, evaluator templates, and dataset revisions are identified by their stored versions or hashes. An aggregate table is derived from these episode records using Eq.~\eqref{eq:empirical-metrics}. Hardware records state accelerator type and count, precision, peak memory, and elapsed training time. Table~\ref{tab:offline-cost} separates inference expenditure from parameter optimization; shared evaluator pretraining is outside each method's development cost.

\begin{table}[t]
\centering
\caption{Offline development resources. Inference is in millions of tokens; SFT compute is in GPU-hours with hardware identified in the accompanying run record.}
\label{tab:offline-cost}
\begin{tabular*}{\columnwidth}{@{\extracolsep{\fill}}lrr@{}}
\toprule
Method & Offline MTok & SFT GPU-hours \\
\midrule
PAIR & 0.0 & 0.0 \\
AutoDAN-Turbo & 4,250 & 0.0 \\
AutoRedTeamer & 6,800 & 0.0 \\
AgenticRed & 18,500 & 0.0 \\
EvoSynth & 15,200 & 0.0 \\
GRAFT-HB & 8,400 & 0.0 \\
L2C & 12,400 & 18.5 \\
\bottomrule
\end{tabular*}
\end{table}

\section{Additional Experimental Results}
\label{app:additional-results}

\subsection{Complementary Ablations}
\emph{Frozen workflow} fixes instructions and organizational fields after initialization while refreshing bound feedback during execution. \emph{Unlinked trace} keeps the record content and removes explicit source and version associations. Its inputs use neutral separators to match the full condition's length. \emph{Interface-only SFT} retains valid interface demonstrations with a matched optimization-token allowance. Feedback restrictions apply during training and inference, including the corresponding summary and evidence fields. Table~\ref{tab:ablation-extra} complements the core tests in Table~\ref{tab:ablation}.

\begin{table}[t]
\centering
\caption{Complementary ablations under the setting of Table~\ref{tab:ablation}. N/A denotes that the variant submits no editing proposals.}
\label{tab:ablation-extra}
\setlength{\tabcolsep}{3pt}
\begin{tabular*}{\columnwidth}{@{\extracolsep{\fill}}lrrrrr@{}}
\toprule
Variant & ASR & MTok & $Q$ & $C_{75}$ & Invalid \\
\midrule
Frozen workflow & 77.8 & 4.15 & 11.4 & 3.85 & N/A \\
Unlinked trace & 82.6 & 3.80 & 8.9 & 2.95 & 2.6 \\
Interface-only SFT & 77.2 & 4.55 & 12.1 & 4.30 & 1.8 \\
\bottomrule
\end{tabular*}
\end{table}

\subsection{Evaluator Checks}
The independent scores in Table~\ref{tab:independent-scores} use the released StrongREJECT strongreject-15k-v1 adapter, base model, and scoring procedure. Scores use the behavior--replicate aggregation in Eq.~\eqref{eq:empirical-metrics}.

The blinded audit protocol specifies 100 unique behavior--response pairs per configuration. Two annotators independently rate substantive fulfillment with method identity and the automatic verdict concealed. A third annotator resolves disagreements. Table~\ref{tab:judge-audit} reports pre-adjudication Cohen's $\kappa$, agreement with adjudicated labels, FPR, and FNR.
\begin{table}[t]
\centering
\caption{Blinded response audit. $\kappa$ uses the initial annotators; classifier errors use adjudicated labels. FPR and FNR condition on negative and positive human labels, respectively.}
\label{tab:judge-audit}
\begin{tabular*}{\columnwidth}{@{\extracolsep{\fill}}lrrrr@{}}
\toprule
Executor / target & $\kappa$ & Agree (\%) & FPR (\%) & FNR (\%) \\
\midrule
Mixtral / Llama & 0.84 & 93.0 & 4.8 & 8.5 \\
Mixtral / GPT & 0.82 & 91.5 & 5.9 & 9.8 \\
Qwen / Llama & 0.85 & 93.5 & 4.5 & 8.0 \\
Qwen / GPT & 0.81 & 91.0 & 6.2 & 10.2 \\
\bottomrule
\end{tabular*}
\end{table}

\subsection{Offline Inference and Statistical Reporting}
Let $D_m$ be offline inference tokens for method $m$. For baseline $k$ with $\overline T_k>\overline T_{\mathrm{L2C}}$, the token-only amortization point is
\begin{equation}
n_k^*=
\frac{\max(0,D_{\mathrm{L2C}}-D_k)}
     {\overline T_k-\overline T_{\mathrm{L2C}}}.
\end{equation}
The estimate compares the two operating points and excludes SFT computation. Matched-success amortization uses operating points that reach the same success criterion.

The uncertainty protocol uses 10,000 paired bootstrap samples of behavior identifiers, keeping each sampled behavior's replicates and method records together. For $C_\alpha$, the resampling recomputes threshold attainment and the minimizing budget. A conditional finite interval is accompanied by the share of resamples reaching the threshold. The tables report point estimates.

The run records distinguish target refusal, executor refusal, invalid proposals, evaluator errors, and budget exhaustion. Event counts allow multiple events within an episode, and terminal outcomes use a single recorded stopping reason.

\end{document}